\documentclass[10pt,twocolumn,letterpaper]{article}

\usepackage[pagenumbers]{cvpr} % To force page numbers, e.g. for an arXiv version

\usepackage[table]{xcolor}
\usepackage{amssymb}
\usepackage{graphicx}
\usepackage{float}

\definecolor{nbarrier}{RGB}{255, 120, 50}
\definecolor{nbicycle}{RGB}{255, 192, 203}
\definecolor{nbus}{RGB}{255, 255, 0}
\definecolor{ncar}{RGB}{0, 150, 245}
\definecolor{nconstruct}{RGB}{0, 255, 255}
\definecolor{nmotor}{RGB}{200, 180, 0}
\definecolor{npedestrian}{RGB}{255, 0, 0}
\definecolor{ntraffic}{RGB}{255, 240, 150}
\definecolor{ntrailer}{RGB}{135, 60, 0}
\definecolor{ntruck}{RGB}{160, 32, 240}
\definecolor{ndriveable}{RGB}{255, 0, 255}
\definecolor{nother}{RGB}{139, 137, 137}
\definecolor{nsidewalk}{RGB}{75, 0, 75}
\definecolor{nterrain}{RGB}{150, 240, 80}
\definecolor{nmanmade}{RGB}{213, 213, 213}
\definecolor{nvegetation}{RGB}{0, 175, 0}

\definecolor{nvcolor}{RGB}{119,185,0}
\definecolor{roadcolor}{RGB}{234,51,246}
\definecolor{sidewalkcolor}{RGB}{68,8,72}
\definecolor{parkingcolor}{RGB}{241,156,249}
\definecolor{othergroundcolor}{RGB}{160,32,76}
\definecolor{buildingcolor}{RGB}{246,202,69}
\definecolor{carcolor}{RGB}{111,149,238}
\definecolor{truckcolor}{RGB}{74,32,172}
\definecolor{bicyclecolor}{RGB}{136,227,242}
\definecolor{motorcyclecolor}{RGB}{37,59,146}
\definecolor{othervehiclecolor}{RGB}{96,81,242}
\definecolor{vegetationcolor}{RGB}{79, 173, 50}
\definecolor{trunkcolor}{RGB}{126, 65, 22}
\definecolor{terraincolor}{RGB}{171, 238, 105}
\definecolor{personcolor}{RGB}{234, 60, 49}
\definecolor{bicyclistcolor}{RGB}{234, 66, 195}
\definecolor{motorcyclistcolor}{RGB}{138, 42, 90}
\definecolor{fencecolor}{RGB}{238, 128, 69}
\definecolor{polecolor}{RGB}{252, 241, 161}
\definecolor{trafficsigncolor}{RGB}{233, 51, 35}
\definecolor{other-struct.color}{RGB}{255, 150, 0}
\definecolor{other-objectcolor}{RGB}{50, 255, 255}
\definecolor{lane-markingcolor}{RGB}{150, 255, 170}
\definecolor{color1}{RGB}{176, 36, 24}
\definecolor{color2}{RGB}{0, 176, 80}
\definecolor{color3}{RGB}{0, 0, 200}

\definecolor{lightbluegray}{RGB}{237,244,250}
\definecolor{tickblue}{RGB}{35,85,135}

\newcommand{\cmark}{\textcolor{tickblue}{\ensuremath{\checkmark}}}

\definecolor{cvprblue}{rgb}{0.21,0.49,0.74}
\usepackage[pagebackref,breaklinks,colorlinks,allcolors=cvprblue]{hyperref}

\def\confName{CVPR}
\def\confYear{2027}

\title{GaussianMamba: Geometric and Relational State Space Scanning of 3D Gaussian Primitives}

\author{A. Enes Doruk\\
{\tt\small abdullahenesdoruk@gmail.com}
\and
Ozgur Erkent\\
Hacettepe University\\
Ankara, Turkiye\\
{\tt\small ozgurerkent@hacettepe.edu.tr}
}

\begin{document}

\maketitle

\begin{abstract}
3D semantic occupancy prediction requires representing complex spatial structures and modeling contextual relationships across a scene. 3D Gaussian primitives provide a sparse and adaptive representation of scene geometry and semantics, while Mamba offers efficient sequence modeling for capturing dependencies among primitives. However, Gaussian primitives are irregular and unordered, making sequence construction challenging, as conventional spatial ordering may overlook their anisotropic geometry and feature-semantic relationships. In this paper, we present GaussianMamba, a dual-branch Mamba framework that explicitly incorporates Gaussian-specific geometric and feature-semantic relationships into complementary scanning strategies. The first branch refines a locality-preserving traversal using anisotropic Gaussian support derived from primitive scale and orientation. The second branch reorders spatially organized groups according to feature-semantic affinity and adjusts local features using geometry-weighted semantic differences. The resulting sequences are processed by separate Mamba modules, and their features are fused through a support-aware mechanism that incorporates local feature discrepancies arising from the distinct scanning paths. Experiments on nuScenes-SurroundOcc and SSCBench-KITTI-360 demonstrate consistent improvements over GaussianFormer and GaussianFormer-2 in both IoU and mIoU.
\end{abstract}    
\section{Introduction}
\label{sec:intro}

\begin{figure}[htb]
    \centering
    \includegraphics[width=0.48\textwidth]{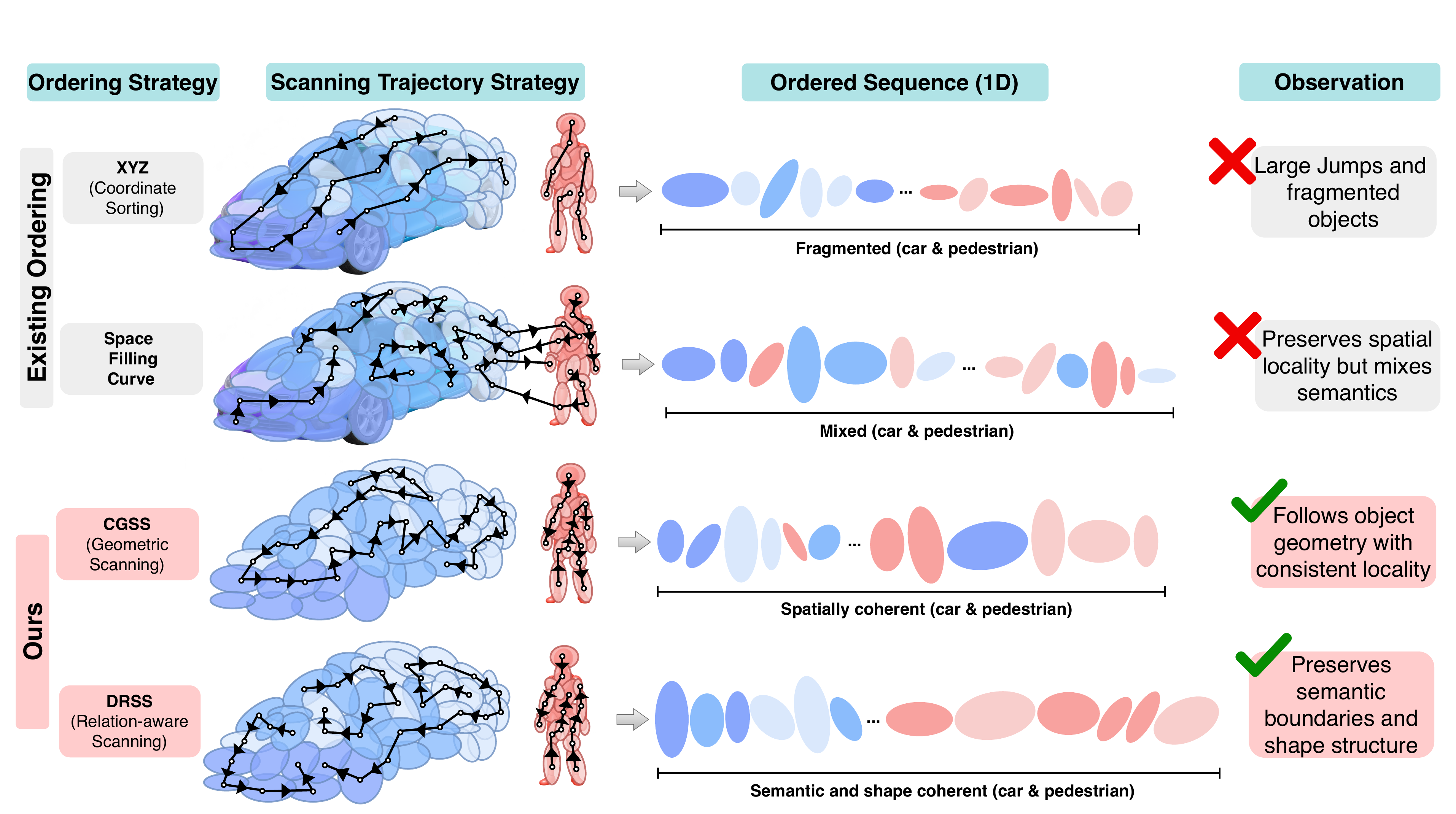}
    \caption{Comparison of Gaussian ordering strategies. Different traversal paths establish different sequential neighborhoods among the same primitives.}
    \label{fig:ordering_comparison}
\end{figure}

3D semantic occupancy prediction jointly reasons about geometry and semantics to produce a dense, scene-level description of foreground objects and background structures, making it useful for autonomous driving~\cite{wei2023surroundocc,wang2023openoccupancy}. Existing methods differ in their spatial representation: BEV is efficient but compresses vertical information, voxels preserve explicit 3D structure at higher cost, and TPV retains spatial information through orthogonal planar projections~\cite{huang2023tpvformer,wang2023openoccupancy}. Recent methods instead represent scenes with sparse 3D Gaussian primitives that carry spatial and semantic attributes~\cite{huang2024gaussianformer,huang2025gaussianformer2}. Regardless of the representation, context exchange among scene elements is typically local: Gaussian refinement relies on sparse convolution and deformable attention over nearby primitives~\cite{huang2024gaussianformer}, whereas dense interaction through self-attention scales quadratically with the number of elements~\cite{occmamba}.

Selective state-space models such as Mamba~\cite{gu2023mamba} provide long-range context at linear cost, but applying them to spatial data requires serializing the elements into scan paths. Prior work traverses image features along multiple directions~\cite{liu2024vmamba}, reorders voxel or planar occupancy features with space-filling or structured traversals~\cite{occmamba,cui2025loma}, and orders points along space-filling curves or coordinate axes~\cite{liang2024pointmamba,zhang2025pcm,wu2024ptv3}. These strategies primarily rely on spatial layouts, point coordinates, or feature-adaptive operations, without exploiting the anisotropic support of Gaussian primitives.

Extending these strategies to 3D Gaussians is less direct. Gaussian primitives form an irregular, unordered set, each defined by a position, anisotropic scale, orientation, and opacity~\cite{huang2024gaussianformer,huang2025gaussianformer2}, so the representation provides no predefined traversal order. Retaining the original primitive indices need not preserve spatial relationships, while coordinate-based sorting may introduce discontinuities and overlook the directional support of anisotropic primitives. Moreover, spatial proximity alone does not capture relationships between learned Gaussian features or their semantic predictions. Since a scan order defines sequential neighborhoods, different orderings yield different adjacency patterns for the same Gaussian set (Fig.~\ref{fig:ordering_comparison}). Existing Gaussian-based Mamba modeling~\cite{doruk2026gaussianocc3d} does not explicitly address support-aware or feature-semantic scan construction. As geometric support and feature-semantic affinity may favor different neighborhoods, a single traversal may not effectively preserve both; this motivates separate scans for the anisotropic geometry and feature-semantic relationships of Gaussian primitives, together with a mechanism to integrate them.

To address these challenges, we introduce GaussianMamba, which incorporates geometry- and relation-aware scanning into the iterative Gaussian refinement architecture for camera-only 3D semantic occupancy prediction. The framework constructs two sequences: one emphasizes spatial continuity and anisotropic Gaussian support, while the other combines feature-semantic group routing with support-weighted boundary correction. Each sequence is processed by its own Mamba branch, and the branch outputs are integrated by a support-aware fusion mechanism that compares how they vary along their respective scan paths. The main contributions are summarized as follows:

\begin{itemize}
    \item We introduce Contact-Guided Spatial Sequencing (CGSS), which incorporates anisotropic Gaussian support into sequence construction through constrained local reordering that accounts for primitive scale and orientation while maintaining spatial continuity.

    \item We design Directional Relation-Aware Semantic Scanning (DRSS), which combines feature-semantic group routing with support-weighted boundary correction to capture relationships beyond geometric adjacency.

    \item We propose Gaussian Support-Discrepancy Fusion (GSDF), which integrates the two scan representations through the support-weighted discrepancy between their scan-local feature variations.
\end{itemize}

\section{Related Work}

\subsection{3D Gaussian Representations for 3D Semantic Occupancy Prediction}

Recent occupancy methods adopt 3D Gaussian primitives as sparse, adaptive alternatives to dense voxel grids. GaussianFormer~\cite{huang2024gaussianformer} represents the scene with semantic Gaussians that are iteratively refined from image features, and GaussianFormer-2~\cite{huang2025gaussianformer2} interprets each Gaussian as a local occupancy distribution whose contributions are combined through probabilistic superposition. For multimodal perception, GaussianFormer3D~\cite{zhao2025gaussianformer3d} incorporates LiDAR priors and 3D deformable attention, Gau-Occ~\cite{lv2026gauocc} uses LiDAR-initialized anchors and geometry-aligned fusion, and GaussianOcc3D~\cite{doruk2026gaussianocc3d} combines LiDAR-aware aggregation with adaptive cross-modal fusion.

Other work exploits relationships among Gaussian primitives. ODG~\cite{shi2025odg} separates static and dynamic components with dual Gaussian query sets, GraphGSOcc~\cite{song2025graphgsocc} models inter-Gaussian dependencies through geometric and semantic graphs whose neighborhoods adapt to Gaussian size and semantic similarity, and GaussianWorld~\cite{zuo2025gaussianworld} propagates Gaussians over time for streaming occupancy. These methods emphasize hierarchical refinement, graph connectivity, or temporal propagation; we instead organize inter-Gaussian relationships as scan orders for selective state-space processing, without constructing explicit graph interactions.

\begin{figure*}[htb]
    \centering
    \includegraphics[width=\textwidth]{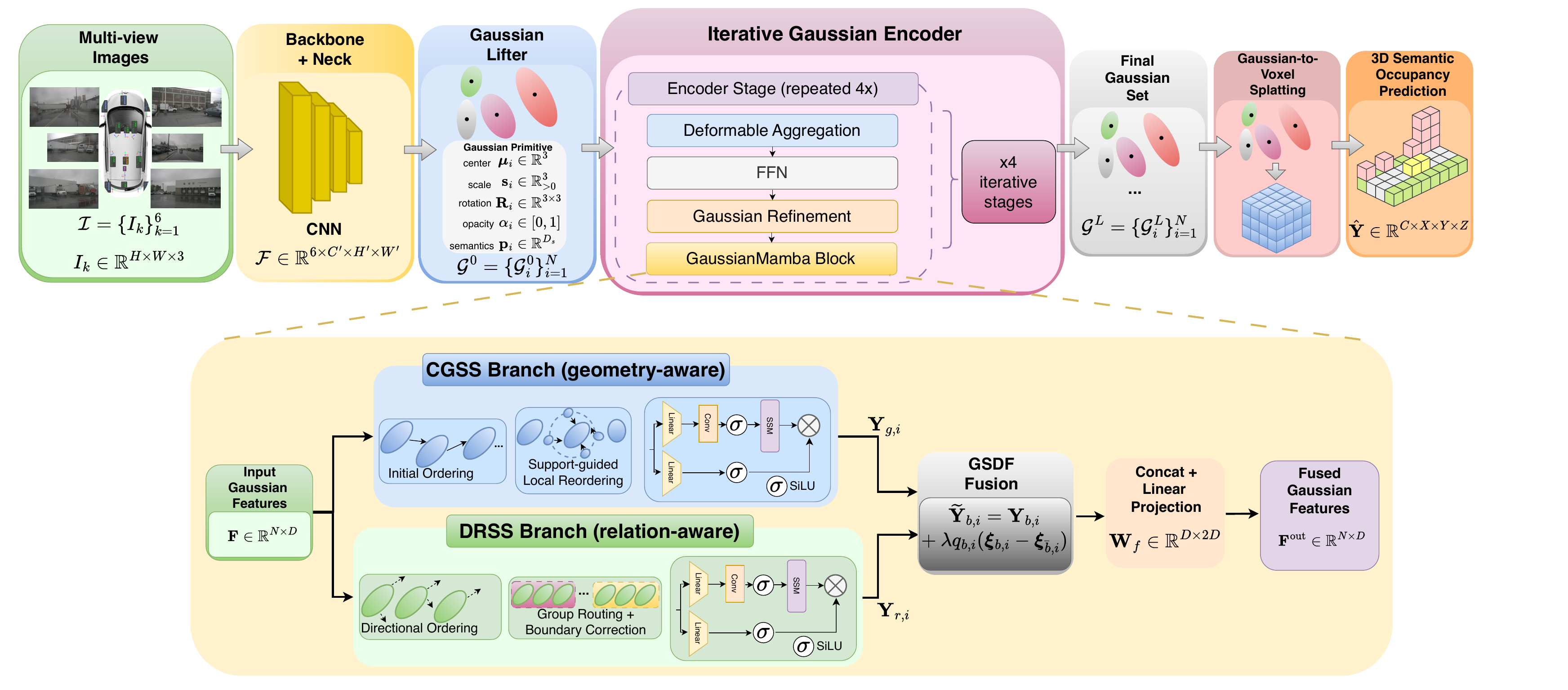}
    \caption{\textbf{Overview of the proposed GaussianMamba framework.}
    Multi-view image features guide iterative Gaussian refinement. CGSS and DRSS construct geometry- and relation-aware scanning sequences, whose Mamba outputs are integrated through Gaussian Support-Discrepancy Fusion (GSDF). The refined Gaussians are splatted into a 3D semantic occupancy grid.}
    \label{fig:framework_overview}
\end{figure*}

\subsection{Mamba-based Occupancy and Spatial Serialization}

Existing Mamba-based occupancy methods differ mainly in how spatial features are serialized. OccMamba~\cite{occmamba} serializes voxel features with a height-prioritized 2D Hilbert expansion and applies hierarchical Mamba blocks for multi-scale context. MambaOcc~\cite{mambaocc} operates on compact BEV features and addresses token-order sensitivity through local adaptive reordering with deformable sampling. FMOcc~\cite{fmocc} applies Plane Selective SSMs to TPV features to suppress redundant empty-space information. GaussianOcc3D~\cite{doruk2026gaussianocc3d} applies a Gauss-Mamba head to Gaussian primitives for state-space context modeling, without explicitly addressing support-aware or feature-semantic scan construction.

Serialization has also been studied for unordered point sets: PointMamba~\cite{liang2024pointmamba} and Point Transformer V3~\cite{wu2024ptv3} order points along space-filling (Hilbert, Z-order) curves, and Point Cloud Mamba~\cite{zhang2025pcm} uses a consistent traverse serialization with axis-permuted variants. Across grid- and point-based methods, scan orders are primarily derived from predefined spatial traversals or feature-space adaptation, whereas Gaussian primitives explicitly parameterize an anisotropic extent and orientation. We employ a shared support-normalized separation measure to guide geometric sequencing, weight feature-semantic boundary correction, and drive discrepancy-based fusion.

\section{Method}
\label{sec:method}

\subsection{Framework Overview}
\label{sec:framework_overview}

Our framework builds upon the Gaussian representation and iterative refinement architecture of GaussianFormer~\cite{huang2024gaussianformer}, incorporating geometry- and relation-aware sequence modeling for camera-only 3D semantic occupancy prediction (Fig.~\ref{fig:framework_overview}). Input images are processed by a dataset-specific CNN backbone and a multi-scale feature neck, whose features guide iterative Gaussian refinement.

We represent a scene using $N$ Gaussian primitives $\{\mathcal{G}_i\}_{i=1}^{N}$, each characterized by its center $\boldsymbol{\mu}_i\in\mathbb{R}^{3}$, anisotropic scale $\mathbf{s}_i\in\mathbb{R}_{>0}^{3}$, rotation $\mathbf{R}_i\in\mathrm{SO}(3)$, stored as a unit quaternion (covariance $\boldsymbol{\Sigma}_i=\mathbf{R}_i\operatorname{diag}(\mathbf{s}_i)^2\mathbf{R}_i^{\top}$), opacity $\alpha_i\in[0,1]$, semantic probability vector $\mathbf{p}_i\in\mathbb{R}^{D_s}$ (the softmax of its current semantic logits), and instance feature $\mathbf{f}_i\in\mathbb{R}^{D}$; the features form $\mathbf{F}\in\mathbb{R}^{N\times D}$.

A state-space model processes its input as a sequence, so the scanning order determines which primitives become neighbors and exchange context most directly. We therefore adapt the scanning layer to Gaussian primitives at three points. Contact-Guided Spatial Sequencing (CGSS) adapts the scanning order using geometry. Directional Relation-Aware Semantic Scanning (DRSS) adapts the order using features and semantics, and corrects the features at predicted semantic boundaries. Gaussian Support-Discrepancy Fusion (GSDF) adapts how the two scans are fused. All three use the same geometric quantity, the support-normalized separation between two Gaussians (Eq.~\eqref{eq:gaussian_support}).

For each branch $b\in\{\mathrm{g},\mathrm{r}\}$ (CGSS and DRSS, respectively), let $\mathcal{P}_b$ apply the scanning permutation $\pi_b$ of that branch. The branch output is
\begin{equation}
\label{eq:branch_operation}
\mathbf{Y}_b
=\mathcal{P}_b^{-1}\!\left[
\mathcal{M}_b\!\left(
\mathcal{P}_b\!\left(
\operatorname{LN}(\mathbf{F})+\boldsymbol{\Delta}_b
\right)
\right)
\right],
\end{equation}
where $\operatorname{LN}$ denotes layer normalization, $\mathcal{M}_b$ an independently parameterized unidirectional Mamba~\cite{gu2023mamba} module, and $\boldsymbol{\Delta}_b\in\mathbb{R}^{N\times D}$ a branch-specific feature correction: CGSS uses none ($\boldsymbol{\Delta}_{\mathrm{g}}=\mathbf{0}$), while $\boldsymbol{\Delta}_{\mathrm{r}}$ introduces support-weighted semantic boundary information. The inverse permutation restores the original Gaussian indices; the permutations themselves receive no gradient. Since the two branches traverse the primitives in different orders, each primitive receives context from two generally different sets of predecessors.

After each refinement stage, a scanning layer consisting of the two Mamba branches and the GSDF module is applied. The refined Gaussians are finally converted into voxel-wise semantic occupancy predictions through Gaussian-to-voxel splatting.

\subsection{Contact-Guided Spatial Sequencing}
\label{sec:cgss}

\begin{figure}[H]
    \includegraphics[width=0.48\textwidth]{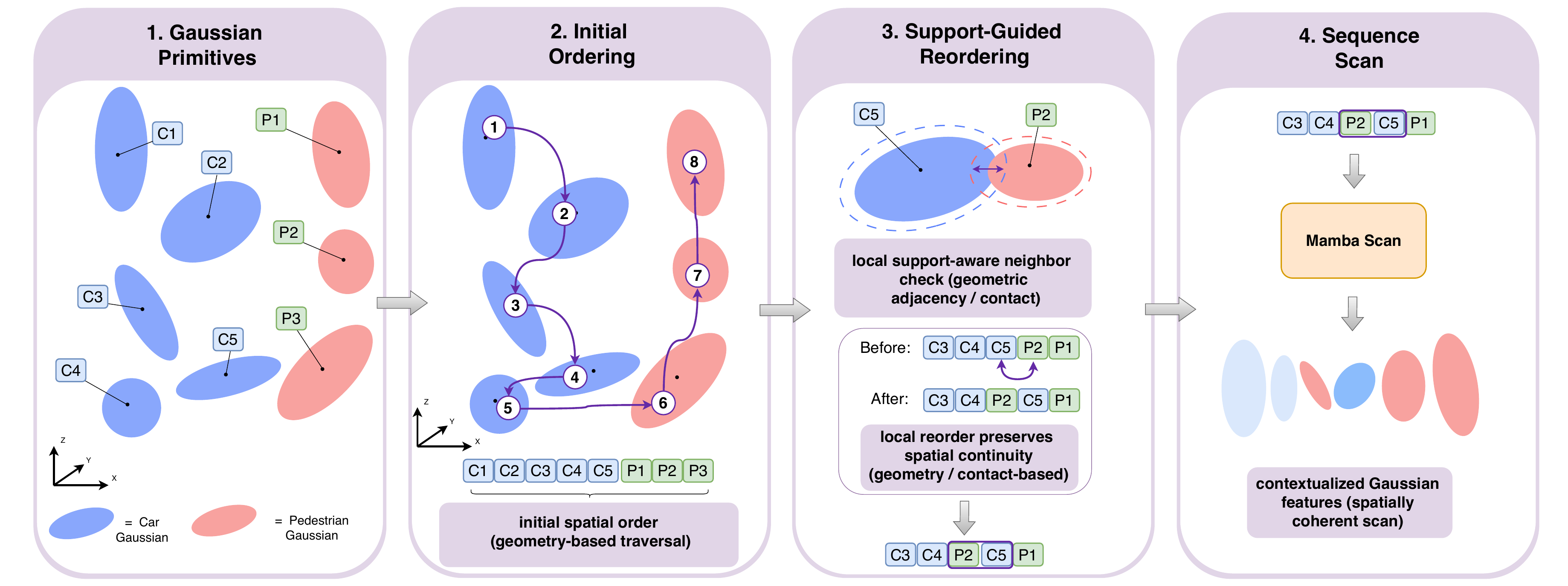}
    \centering
    \caption{Contact-Guided Spatial Sequencing (CGSS). Local reordering incorporates anisotropic Gaussian support while limiting deviations from the initial spatial traversal, without enforcing semantic grouping.}
    \label{fig:cgss}
\end{figure}

Gaussian primitives have no inherent ordering. A spatial traversal based on Gaussian centers provides an initial organization, but center proximity alone does not fully characterize the relationships between anisotropic primitives. In particular, at the same center distance, two Gaussians may overlap or be clearly separated, depending on their scales and orientations. To account for these properties, we introduce Contact-Guided Spatial Sequencing (CGSS, Fig.~\ref{fig:cgss}), which constructs Gaussian-specific geometric adjacency while maintaining the broader spatial organization of the sequence.

We first quantize the Gaussian centers and construct an initial traversal using a 3D Hilbert space-filling curve, whose spatial locality provides a suitable initialization for neighborhood-based refinement. Although directional Gaussian support (defined below) offers an additional criterion for determining adjacency, optimizing this measure alone does not ensure spatial continuity across the sequence. CGSS therefore retains the initial traversal as a spatial reference and introduces constrained local refinement.

For two Gaussians $\mathcal{G}_i$ and $\mathcal{G}_j$, we define their  center distance as $d_{ij}=\|\boldsymbol{\mu}_j-\boldsymbol{\mu}_i\|_2$ and their normalized displacement direction as $\mathbf{n}_{ij}=(\boldsymbol{\mu}_j-\boldsymbol{\mu}_i)/(d_{ij}+\epsilon)$, with a small constant $\epsilon$. With the rotation matrix $\mathbf{R}_i$, the directional support of Gaussian $i$ and the support-normalized separation between the two primitives are computed as
\begin{equation}
\label{eq:gaussian_support}
\begin{aligned}
\rho_i(\mathbf{n}_{ij})
&=\left\|
\operatorname{diag}(\mathbf{s}_i)
\mathbf{R}_i^{\top}\mathbf{n}_{ij}
\right\|_2,\\
g_{ij}
&=\frac{d_{ij}}
{\rho_i(\mathbf{n}_{ij})+\rho_j(\mathbf{n}_{ij})+\epsilon}.
\end{aligned}
\end{equation}
The scalar $\rho_i$ is the standard deviation of Gaussian $i$ projected onto $\mathbf{n}_{ij}$, i.e., its one-standard-deviation extent along that direction. Consequently, $g_{ij}\in\mathbb{R}_{\geq0}$ measures center separation relative to the combined directional supports of the two primitives: $g_{ij}\leq1$ means that the one-standard-deviation extents of the two Gaussians overlap when projected onto the line connecting their centers, which we use as a proxy for contact, whereas $g_{ij} > 1$ implies that their one-standard-deviation ellipsoids are disjoint. It enters the three components through different monotone mappings (Eqs.~\eqref{eq:cgss_adjacency}, \eqref{eq:boundary_weights}, and \eqref{eq:gsdf_support}).

To incorporate this measure while limiting deviations from the initial spatial traversal, we define an adjacency cost:
\begin{equation}
\label{eq:cgss_adjacency}
c_{ij}
=\log(1+g_{ij})
+\frac{d_{ij}}{d_0}
+\frac{|h_i-h_j|}{W},
\end{equation}
where $c_{ij}\geq0$ is the adjacency cost, $h_i$ and $h_j$ denote the initial sequence ranks, $W$ is the local refinement window size, and $d_0$ is a spatial normalization factor. The first term favors pairs with a small support-normalized separation, while the normalized distance and rank terms penalize large spatial transitions and adjacency between primitives that were distant in the initial traversal.

The sequence is refined through adjacent swaps within non-overlapping local windows of $W$ primitives, whose endpoints are kept fixed, so no primitive leaves its window. Each candidate swap is evaluated by comparing the sum of adjacency costs over the affected neighboring pairs. A swap is accepted only if it decreases this cost while satisfying constraints on both the total and maximum affected edge lengths, which restrict changes in spatial connectivity. After a fixed number of local sweeps, the resulting permutation $\pi_{\mathrm{g}}$ determines the feature sequence processed by the CGSS Mamba module.

\subsection{Directional Relation-Aware Semantic Scanning}
\label{sec:drss}

\begin{figure}[hbt]
    \includegraphics[width=0.49\textwidth]{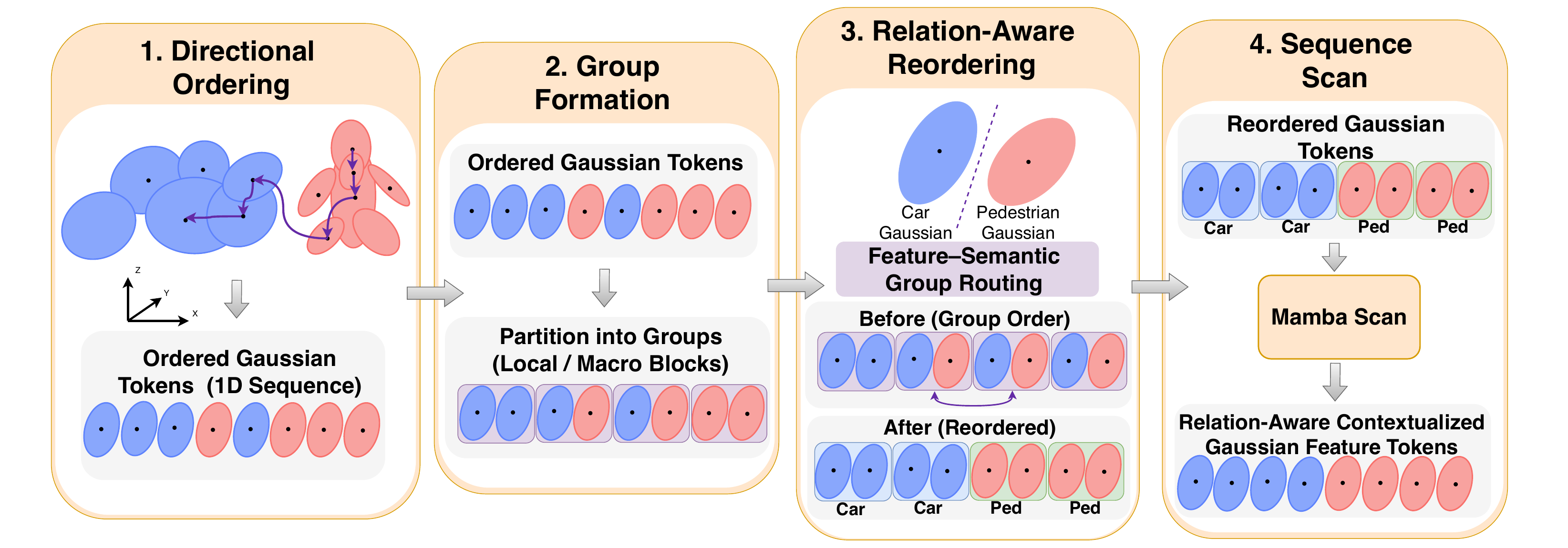}
    \centering
    \caption{Directional Relation-Aware Semantic Scanning (DRSS). Feature–semantic affinity guides local group routing, while support-weighted boundary correction refines Gaussian features.}
    \label{fig:drss}
\end{figure}

CGSS establishes Gaussian adjacency through directional support, but this criterion does not explicitly capture relationships between learned instance features or predicted semantic distributions. Moreover, Gaussian feature relationships can be considered at different levels of granularity: group-level associations describe relationships across spatial regions, whereas interactions between individual primitives characterize local semantic differences. To incorporate these complementary relationships, we introduce Directional Relation-Aware Semantic Scanning (DRSS, Fig.~\ref{fig:drss}), which combines relation-aware group routing with support-weighted semantic boundary correction while retaining a structured spatial traversal.

\begin{table*}[htb] %
    \caption{Surround view 3D semantic occupancy prediction results on nuScenes-SurroundOcc. TPVFormer* is supervised by dense occupancy annotations. }
    \small
    \setlength{\tabcolsep}{0.005\linewidth}  
    \renewcommand\arraystretch{1.05}
    \centering
    \resizebox{\textwidth}{!}{
    \begin{tabular}{l|c c | c c c c c c c c c c c c c c c c}
        \toprule
        Method
        & IoU
        & mIoU
        & \rotatebox{90}{\textcolor{nbarrier}{$\blacksquare$} barrier}
        & \rotatebox{90}{\textcolor{nbicycle}{$\blacksquare$} bicycle}
        & \rotatebox{90}{\textcolor{nbus}{$\blacksquare$} bus}
        & \rotatebox{90}{\textcolor{ncar}{$\blacksquare$} car}
        & \rotatebox{90}{\textcolor{nconstruct}{$\blacksquare$} const. veh.}
        & \rotatebox{90}{\textcolor{nmotor}{$\blacksquare$} motorcycle}
        & \rotatebox{90}{\textcolor{npedestrian}{$\blacksquare$} pedestrian}
        & \rotatebox{90}{\textcolor{ntraffic}{$\blacksquare$} traffic cone}
        & \rotatebox{90}{\textcolor{ntrailer}{$\blacksquare$} trailer}
        & \rotatebox{90}{\textcolor{ntruck}{$\blacksquare$} truck}
        & \rotatebox{90}{\textcolor{ndriveable}{$\blacksquare$} drive. suf.}
        & \rotatebox{90}{\textcolor{nother}{$\blacksquare$} other flat}
        & \rotatebox{90}{\textcolor{nsidewalk}{$\blacksquare$} sidewalk}
        & \rotatebox{90}{\textcolor{nterrain}{$\blacksquare$} terrain}
        & \rotatebox{90}{\textcolor{nmanmade}{$\blacksquare$} manmade}
        & \rotatebox{90}{\textcolor{nvegetation}{$\blacksquare$} vegetation}
        \\
        \midrule

BEVFormer~\cite{li2024bevformer}
& 30.50 & 16.75 & 14.22 & 6.58 & 23.46 & 28.28
& 8.66 & 10.77 & 6.64 & 4.05 & 11.20 & 17.78
& 37.28 & 18.00 & 22.88 & 22.17 & 13.80 & \textbf{22.21} \\

TPVFormer~\cite{huang2023tpvformer}
& 11.51 & 11.66 & 16.14 & 7.17 & 22.63 & 17.13
& 8.83 & 11.39 & 10.46 & 8.23 & 9.43 & 17.02
& 8.07 & 13.64 & 13.85 & 10.34 & 4.90 & 7.37 \\

TPVFormer*~\cite{huang2023tpvformer}
& 30.86 & 17.10 & 15.96 & 5.31 & 23.86 & 27.32
& 9.79 & 8.74 & 7.09 & 5.20 & 10.97 & 19.22
& 38.87 & 21.25 & 24.26 & 23.15 & 11.73 & 20.81 \\

OccFormer~\cite{zhang2023occformer}
& 31.39 & 19.03 & 18.65 & 10.41 & 23.92 & 30.29
& 10.31 & 14.19 & 13.59 & 10.13 & 12.49 & 20.77
& 38.78 & 19.79 & 24.19 & 22.21 & 13.48 & 21.35 \\

SurroundOcc~\cite{wei2023surroundocc}
& 31.49 & 20.30 & \textbf{20.59} & 11.68
& 28.06 & 30.86 & 10.70 & 15.14
& \textbf{14.09} & \textbf{12.06} & \textbf{14.38}
& 22.26 & 37.29 & 23.70 & 24.49 & 22.77
& \textbf{14.89} & 21.86 \\

\midrule

GaussianFormer~\cite{huang2024gaussianformer}
& 29.83 & 19.10 & 19.52 & 11.26 & 26.11 & 29.78
& 10.47 & 13.83 & 12.58 & 8.67 & 12.74 & 21.57
& 39.63 & 23.28 & 24.46 & 22.99 & 9.59 & 19.12 \\

\rowcolor{lightbluegray}
\textbf{GaussianFormer + Ours}$^\dagger$
& 32.41 & 20.55 & 19.36 & 14.04 & 26.32
& \textbf{32.10} & 8.46 & 15.96 & 13.34 & 9.27
& 12.57 & 22.01 & \textbf{45.88} & \textbf{26.04}
& \textbf{27.72} & \textbf{25.72} & 10.18 & 19.76 \\

\midrule

GaussianFormer-2~\cite{huang2025gaussianformer2}
& 30.56 & 20.02 & 20.15 & 12.99 & 27.61 & 30.23
& 11.19 & 15.31 & 12.64 & 9.63 & 13.31 & 22.26
& 39.68 & 23.47 & 25.62 & 23.20 & 12.25 & 20.73 \\

\rowcolor{lightbluegray}
\textbf{GaussianFormer-2 + Ours}$^\dagger$
& \textbf{32.98} & \textbf{21.12} & 20.27
& \textbf{14.25} & \textbf{29.45} & 31.64
& \textbf{11.92} & \textbf{16.65} & 13.04
& 10.45 & 13.95 & \textbf{23.03} & 42.82
& 23.50 & 26.82 & 25.59 & 12.75 & 21.86 \\
        \bottomrule
    \end{tabular}}
    \label{tab: nuscenes results}
\end{table*}

\textbf{Relation-aware group routing.} We first construct a directional (XYZ) spatial traversal from the quantized Gaussian centers. The traversal prioritizes the $x$-axis, alternating the $y$ direction between consecutive $x$ slices and the $z$ direction between traversed $y$ rows; sorting the primitives along this traversal establishes the initial serpentine sequence. This spatial organization provides a consistent basis for grouping primitives and avoids an unrestricted global ordering driven solely by semantic predictions.

The ordered primitives are partitioned into groups of $S$ consecutive Gaussians, with $K$ groups forming a macro-block. Each group $u$ is represented by an $\ell_2$-normalized mean instance feature $\bar{\mathbf{f}}_u\in\mathbb{R}^{D}$, an $\ell_2$-normalized semantic descriptor $\bar{\mathbf{p}}_u\in\mathbb{R}^{D_s}$, and a mean spatial center $\bar{\boldsymbol{\mu}}_u\in\mathbb{R}^{3}$. The semantic descriptor aggregates predicted class probabilities from confident Gaussians only (maximum class probability $\geq\tau_c$), reducing the contribution of uncertain predictions to group relationships.

To reorganize groups according to both their represented content and spatial structure, we define the pairwise affinity
\begin{equation}
\label{eq:group_affinity}
\begin{aligned}
A_{uv}
={}&\bar{\mathbf{f}}_u^{\top}\bar{\mathbf{f}}_v
+m_{uv}\bar{\mathbf{p}}_u^{\top}\bar{\mathbf{p}}_v\\
&-\min\!\left(
\frac{\|\bar{\boldsymbol{\mu}}_u-\bar{\boldsymbol{\mu}}_v\|_2}{d_a},3
\right) -\frac{|u-v|}{K-1},
\end{aligned}
\end{equation}
where $A_{uv}\in\mathbb{R}$ measures the affinity between groups $u$ and $v$ of the same macro-block, $m_{uv}\in\{0,1\}$ is a semantic reliability gate, and $d_a$ is a distance normalization factor. The descriptor similarities capture feature and semantic agreement, while the bounded spatial-distance term and normalized rank difference keep the routing spatially conservative. The semantic term is disabled ($m_{uv}=0$) when either group has insufficient confident predictions or when the scene-level predicted class distribution is overwhelmingly concentrated in a single class.

Within each macro-block, routing begins from the first group and greedily selects the highest-affinity unvisited group until all interior groups have been assigned. The final group remains fixed, preserving the macro-block endpoints. Group-internal Gaussian ordering and the sequence of macro-blocks are also retained; primitives outside complete macro-blocks keep their initial serpentine order.

\textbf{Support-weighted boundary correction.} While group-level routing models relationships through aggregated descriptors, such descriptors do not explicitly encode semantic differences between individual neighboring primitives. To complement this group-level organization, we introduce a support-weighted semantic boundary correction that modifies individual Gaussian features rather than their ordering. Candidate neighbors are obtained from the Hilbert traversal used to initialize CGSS, which serves here only to find spatially close pairs. A fixed number of preceding and following candidates are examined, retaining only pairs satisfying spatial distance and semantic confidence constraints.

For an eligible pair $(i,j)$, semantic disagreement is measured using the total-variation distance $t_{ij}=\frac12\|\mathbf{p}_i-\mathbf{p}_j\|_1$, where $t_{ij}\in[0,1]$. To incorporate the geometric relationship between these primitives, we combine semantic disagreement with spatial proximity, opacity, and the support-normalized separation $g_{ij}$ of Eq.~\eqref{eq:gaussian_support}:
\begin{equation}
\label{eq:boundary_weights}
\begin{aligned}
\eta_{ij}
&=\sqrt{\alpha_i\alpha_j}\exp\!\left[
-\frac12\bigl(\max(g_{ij}-1,0)\bigr)^2
\right],\\
w_{ij}
&=t_{ij}\exp\!\left[-\left(\frac{d_{ij}}{r_0}\right)^2\right]
\hat{\eta}_{ij}.
\end{aligned}
\end{equation}
Here, $\eta_{ij}\in[0,1]$ denotes the Gaussian contact factor, which decays as the support-normalized separation exceeds one, and $r_0$ is the boundary neighborhood radius. The normalized contact factor $\hat{\eta}_{ij}$ is $\eta_{ij}$ divided by its weighted scene-level average and clipped. The resulting score $w_{ij}\geq0$ is therefore large when two nearby Gaussians in contact are predicted to belong to different classes, i.e., at a likely semantic boundary. The score is treated as a constant; gradients flow only through the feature difference in Eq.~\ref{eq:boundary_correction}.

For each Gaussian $i$, we select the eligible neighbor $j^*(i)$ with the highest boundary score and compute
\begin{equation}
\label{eq:boundary_correction}
\boldsymbol{\Delta}_{\mathrm{r},i}
=w_{i,j^*(i)}\left[
\operatorname{LN}(\mathbf{f}_i)-
\operatorname{LN}(\mathbf{f}_{j^*(i)})
\right],
\end{equation}
where $\boldsymbol{\Delta}_{\mathrm{r},i}\in\mathbb{R}^{D}$ is the boundary-sensitive feature correction for primitive $i$. That is, the scan input of $i$ is moved away from the normalized feature of its neighbor across the predicted boundary, in proportion to the boundary score. The correction is zero if no eligible neighbor is found, and it is added to the normalized Gaussian features before the relation-aware permutation $\pi_{\mathrm{r}}$ is applied (Eq.~\eqref{eq:branch_operation}).

\subsection{Gaussian Support-Discrepancy Fusion}
\label{sec:gsdf}

CGSS and DRSS establish different neighborhood relationships among the same Gaussian primitives through geometry- and relation-aware scanning, respectively. Concatenation integrates the two outputs of each primitive but does not explicitly compare how each output varies among its sequence neighbors. To account for these differences during feature integration, we introduce Gaussian Support-Discrepancy Fusion (GSDF), which incorporates scan-dependent feature discrepancies and anisotropic Gaussian support.

Let $\mathbf{Y}_{\mathrm{g}}$ and $\mathbf{Y}_{\mathrm{r}}$ denote the outputs of CGSS and DRSS, respectively, after restoring the original Gaussian indexing. For branch $b$, $b'$ denotes the other branch. For each Gaussian $i$, $\mathcal{N}_b(i)$ contains its immediately preceding and following primitives in the scanning order of branch $b$, when available. Using the support-normalized separation $g_{ij}$ defined in Eq.~\eqref{eq:gaussian_support}, we define the Gaussian support affinity as $k_{ij}=\exp(-g_{ij})$ and compute
\begin{equation}
\label{eq:gsdf_support}
\begin{aligned}
\boldsymbol{\xi}_{b,i}
&=
\frac{
\sum_{j\in\mathcal{N}_b(i)}
k_{ij}(\mathbf{Y}_{b,j}-\mathbf{Y}_{b,i})
}{
1+\sum_{j\in\mathcal{N}_b(i)}k_{ij}
},\\
q_{b,i}
&=
\frac{1}{|\mathcal{N}_b(i)|}
\sum_{j\in\mathcal{N}_b(i)}k_{ij}.
\end{aligned}
\end{equation}
Here, $\boldsymbol{\xi}_{b,i}\in\mathbb{R}^{D}$ captures support-weighted local feature variation along each scanning path, while $q_{b,i}$ down-weights the adjustment in Eq.~\eqref{eq:gsdf_fusion} where the scan neighbors are far apart relative to their supports.

The difference between the branch-specific feature contrasts is incorporated into the Gaussian representations before their integration:
\begin{equation}
\label{eq:gsdf_fusion}
\begin{aligned}
\widetilde{\mathbf{Y}}_{b,i}
&=\mathbf{Y}_{b,i}
+\lambda\,q_{b,i}
(\boldsymbol{\xi}_{b,i}-\boldsymbol{\xi}_{b',i}),\\
\mathbf{f}_i^{\mathrm{out}}
&=\mathbf{W}_f
[\widetilde{\mathbf{Y}}_{\mathrm{g},i};
\widetilde{\mathbf{Y}}_{\mathrm{r},i}]
+\mathbf{b}_f.
\end{aligned}
\end{equation}
Here, $\mathbf{W}_f\in\mathbb{R}^{D\times 2D}$ and $\mathbf{b}_f\in\mathbb{R}^{D}$ denote the learned projection parameters, and $\lambda=\tanh(\theta)$ is a learnable scalar initialized to zero, so that GSDF starts from plain concatenation fusion. Equivalently, the difference between the two adjusted branches changes by $\lambda(q_{\mathrm{g},i}+q_{\mathrm{r},i})(\boldsymbol{\xi}_{\mathrm{g},i}-\boldsymbol{\xi}_{\mathrm{r},i})$: GSDF amplifies or suppresses the inter-branch difference where the two scans see different local variation. The fused feature $\mathbf{f}_i^{\mathrm{out}}$ replaces $\mathbf{f}_i$ as the input to the next block.

\begin{table*}[htb]
    \centering
    \caption{Monocular 3D semantic occupancy prediction results on SSCBench-KITTI-360.}
    \label{tab:kittiseg}
    \setlength{\tabcolsep}{0.005\linewidth}
    \renewcommand\arraystretch{1.05}
    \resizebox{\linewidth}{!}{
    \begin{tabular}{l|c|c|c|cccccccccccccccccc}
    \toprule
    Method
    & \rotatebox{90}{Input}
    & IoU
    & mIoU
    & \rotatebox{90}{\textcolor{carcolor}{$\blacksquare$} car}
    & \rotatebox{90}{\textcolor{bicyclecolor}{$\blacksquare$} bicycle}
    & \rotatebox{90}{\textcolor{motorcyclecolor}{$\blacksquare$} motorcycle}
    & \rotatebox{90}{\textcolor{truckcolor}{$\blacksquare$} truck}
    & \rotatebox{90}{\textcolor{othervehiclecolor}{$\blacksquare$} other-veh.}
    & \rotatebox{90}{\textcolor{personcolor}{$\blacksquare$} person}
    & \rotatebox{90}{\textcolor{roadcolor}{$\blacksquare$} road}
    & \rotatebox{90}{\textcolor{parkingcolor}{$\blacksquare$} parking}
    & \rotatebox{90}{\textcolor{sidewalkcolor}{$\blacksquare$} sidewalk}
    & \rotatebox{90}{\textcolor{othergroundcolor}{$\blacksquare$} other-grnd}
    & \rotatebox{90}{\textcolor{buildingcolor}{$\blacksquare$} building}
    & \rotatebox{90}{\textcolor{fencecolor}{$\blacksquare$} fence}
    & \rotatebox{90}{\textcolor{vegetationcolor}{$\blacksquare$} vegetation}
    & \rotatebox{90}{\textcolor{terraincolor}{$\blacksquare$} terrain}
    & \rotatebox{90}{\textcolor{polecolor}{$\blacksquare$} pole}
    & \rotatebox{90}{\textcolor{trafficsigncolor}{$\blacksquare$} traf.-sign}
    & \rotatebox{90}{\textcolor{other-struct.color}{$\blacksquare$} other-struct.}
    & \rotatebox{90}{\textcolor{other-objectcolor}{$\blacksquare$} other-object}
    \\
    \midrule

   MonoScene~\cite{cao2022monoscene}
    & C & 37.87 & 12.31
    & 19.34 & 0.43 & 0.58 & 8.02
    & 2.03 & 0.86 & 48.35 & 11.38
    & 28.13 & 3.22 & 32.89 & 3.53
    & 26.15 & 16.75 & 6.92 & 5.67
    & 4.20 & 3.09 \\
    
    Voxformer~\cite{li2023voxformer}
    & C & 38.76 & 11.91
    & 17.84 & 1.16 & 0.89 & 4.56
    & 2.06 & 1.63 & 47.01 & 9.67
    & 27.21 & 2.89 & 31.18 & 4.97
    & 28.99 & 14.69 & 6.51 & 6.92
    & 3.79 & 2.43 \\
    
    TPVFormer~\cite{huang2023tpvformer}
    & C & 40.22 & 13.64
    & 21.56 & 1.09 & 1.37 & 8.06
    & 2.57 & 2.38 & 52.99 & 11.99
    & 31.07 & 3.78 & 34.83 & 4.80
    & 30.08 & 17.51 & 7.46 & 5.86
    & 5.48 & 2.70 \\
    
    OccFormer~\cite{zhang2023occformer}
    & C & 40.27 & 13.81
    & 22.58 & 0.66 & 0.26 & 9.89
    & 3.82 & 2.77 & 54.30 & \textbf{13.44}
    & 31.53 & 3.55 & \textbf{36.42} & 4.80
    & 31.00 & \textbf{19.51} & \textbf{7.77}
    & \textbf{8.51} & 6.95 & 4.60 \\
    
    \midrule
    
    GaussianFormer~\cite{huang2024gaussianformer}
    & C & 35.38 & 12.92
    & 18.93 & 1.02 & 4.62 & 18.07
    & \textbf{7.59} & 3.35 & 45.47 & 10.89
    & 25.03 & \textbf{5.32} & 28.44 & 5.68
    & 29.54 & 8.62 & 2.99 & 2.32
    & 9.51 & \textbf{5.14} \\
    
    \rowcolor{lightbluegray}
    \textbf{GaussianFormer + Ours}$^\dagger$
    & C & 37.87 & 13.95
    & 18.81 & 1.32 & \textbf{5.02} & \textbf{20.17}
    & 7.34 & \textbf{3.80} & 49.29 & 12.20
    & 27.87 & 5.14 & 31.27 & \textbf{6.32}
    & 31.95 & 8.30 & 3.39 & 2.82
    & \textbf{11.06} & 5.03 \\
    
    \midrule
    
    GaussianFormer-2~\cite{huang2025gaussianformer2}
    & C & 38.37 & 13.90
    & 21.08 & 2.55 & 4.21 & 12.41
    & 5.73 & 1.59 & 54.12 & 11.04
    & 32.31 & 3.34 & 32.01 & 4.98
    & 28.94 & 17.33 & 3.57 & 5.48
    & 5.88 & 3.54 \\
    
    \rowcolor{lightbluegray}
    \textbf{GaussianFormer-2 + Ours}$^\dagger$
    & C & \textbf{40.57} & \textbf{14.80}
    & \textbf{23.08} & \textbf{2.79}
    & 4.07 & 13.31
    & 5.52 & 1.85 & \textbf{56.36} & 12.60
    & \textbf{35.38} & 3.16 & 34.68 & 5.43
    & \textbf{32.26} & 16.98 & 4.12 & 5.23
    & 5.70 & 3.88 \\

    \bottomrule
    \end{tabular}
    }
\end{table*}

\section{Experiment}

\subsection{Dataset and Metric}

\textbf{The nuScenes-SurroundOcc dataset}~\cite{caesar2020nuscenes} contains 1K surround-view driving scenes collected in Boston and Singapore, officially split into 700 training, 150 validation, and 150 testing scenes. Each scene spans 20 seconds with annotations at 2 Hz and includes camera, LiDAR, radar, and IMU data. We adopt the 3D semantic occupancy annotations from SurroundOcc~\cite{wei2023surroundocc} for supervision and evaluation. The annotated volume covers $[-50,50]$ m along the x- and y-axes and $[-5,3]$ m along the z-axis, with a voxel size of 0.5 m. The 18 occupancy labels comprise 16 semantic categories, one empty class, and one noise class.

\textbf{The KITTI-360 dataset}~\cite{liao2022kitti} provides over 320K images of suburban driving environments, captured using multiple cameras and LiDAR sensors. We use only the ego vehicle's left-camera images as input. For occupancy supervision and evaluation, we adopt SSCBench-KITTI-360~\cite{li2024sscbench}, officially split into 7 training, 1 validation, and 1 testing sequence, containing 8,487, 1,812, and 2,566 keyframes, respectively. The annotated region spans $51.2\times51.2\times6.4$ $\mathrm{m}^3$ ahead of the vehicle, discretized into $256\times256\times32$ voxels with 19 labels, including 18 semantic classes and one empty class.

\textbf{The evaluation metrics} adhere to common practice, namely mean Intersection-over-Union (mIoU) and Intersection-over-Union (IoU):

\begin{equation}
\mathbf{mIoU} = \frac{1}{|\mathcal{C}'|}\sum_{i\in \mathcal{C}'}{\frac{TP_i}{TP_i+FP_i+FN_i}},
\end{equation}
\begin{equation}
\mathbf{IoU} = \frac{TP_{\neq c_0}}{TP_{\neq c_0}+FP_{\neq c_0}+FN_{\neq c_0}},
\end{equation}
where $\mathcal{C}'$ is the set of non-empty classes, $c_0$ is the empty class, and TP, FP, and FN count the true-positive, false-positive, and false-negative voxels.

\subsection{Implementation Details}
For nuScenes-SurroundOcc~\cite{caesar2020nuscenes} and KITTI-360~\cite{liao2022kitti}, we use input image resolutions of $900\times1600$ and $376\times1408$. Following GaussianFormer~\cite{huang2024gaussianformer}, we initialize the image backbones using the same pretrained checkpoints: ResNet101-DCN~\cite{resnet} with FCOS3D pretrained weights~\cite{wang2021fcos3d} for nuScenes-SurroundOcc, and ImageNet-pretrained~\cite{deng2009imagenet} ResNet50~\cite{resnet} for KITTI-360. On nuScenes-SurroundOcc, GaussianFormer uses 144K Gaussian primitives, whereas GaussianFormer-2 employs 12.8K primitives with 128-dimensional features. For KITTI-360, we use 38.4K Gaussians. The scanning layers use $W{=}32$, $d_0{=}3$\,m, $S{=}128$, $K{=}8$, $\tau_c{=}0.16$, $d_a{=}10$\,m, and $r_0{=}2$\,m (two sweeps; total/maximum edge growth $\leq$5\%/10\%; gate: $\geq S/8$ confident Gaussians per group, dominant class $<$90\%; six candidates per side). We train with the unchanged cross-entropy and Lov\'asz-softmax~\cite{berman2018lovasz} losses of GaussianFormer and GaussianFormer-2~\cite{huang2024gaussianformer,huang2025gaussianformer2} using AdamW~\cite{adamw} with a weight decay of 0.01 and a peak learning rate of $2\times10^{-4}$, following a cosine annealing schedule. Training is conducted for 20 epochs with a batch size of 8 on nuScenes-SurroundOcc and 30 epochs with a batch size of 4 on KITTI-360.

\subsection{Results}

\begin{figure*}[htb]
    \centering
    \includegraphics[width=0.8\textwidth]{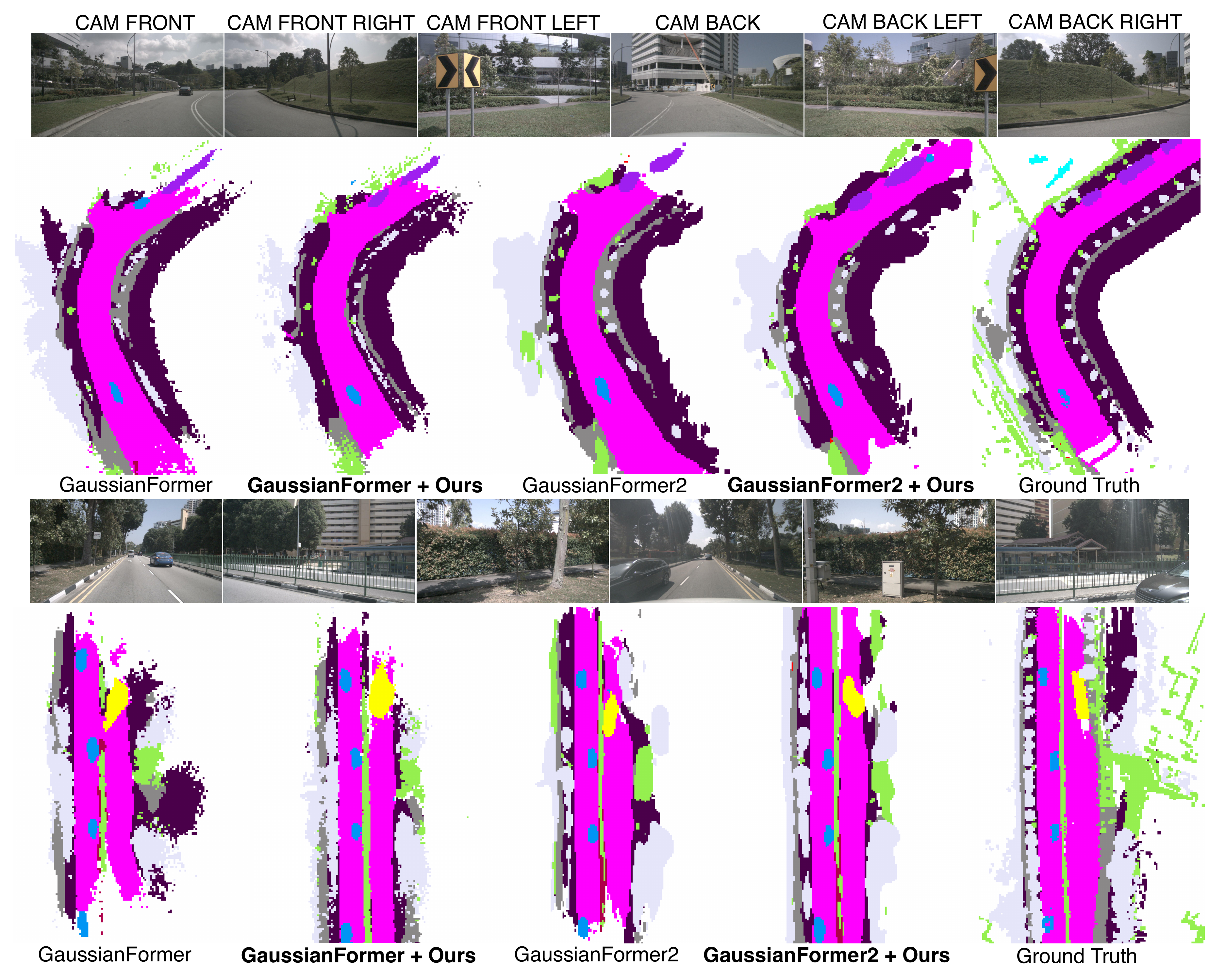}
\caption{Qualitative comparison of GaussianFormer and GaussianFormer-2 with and without GaussianMamba on nuScenes-SurroundOcc.}  
\label{fig:qual}
\end{figure*}

\definecolor{lightbluegray}{RGB}{237,244,250}

\#B and \#G denote Mamba branches and Gaussian counts; $^\dagger$ indicates three-seed accuracy averages. All ablations use 144K Gaussians. Dual branches use independent Mamba parameters, defaulting to Hilbert scanning and concatenation fusion unless specified otherwise. Latency is measured on an NVIDIA RTX 6000 Ada and reported per scanning layer, which is applied after each of the four refinement stages; ordering ablations cover scanning and Mamba processing, while fusion parameters and latency are module-specific.

\noindent\textbf{Results on nuScenes-SurroundOcc.}
Table~\ref{tab: nuscenes results} shows that GaussianMamba improves GaussianFormer by 2.58 IoU and 1.45 mIoU points, surpassing SurroundOcc, while GaussianFormer-2 gains 2.42 and 1.10 points, achieving the highest IoU and mIoU in the comparison. Improvements span all 16 classes with GaussianFormer-2 and 13 with GaussianFormer.
Notably, drivable surface gains 6.25 points with GaussianFormer and 3.14 with GaussianFormer-2, with consistent improvements in terrain, suggesting benefits for spatially extended regions.

\noindent\textbf{Results on SSCBench-KITTI-360.}
For monocular occupancy prediction (Table~\ref{tab:kittiseg}), GaussianMamba improves GaussianFormer by 2.49 IoU and 1.03 mIoU points,
and GaussianFormer-2 by 2.20 and 0.90 points, the latter achieving the highest overall scores. Gains are particularly evident in spatially extended classes, including road, sidewalk, building, and vegetation (+2.2 to +3.8 points), while individual class regressions remain within 0.35 points. These results suggest consistent benefits across different Gaussian representations and benchmarks.

\begin{table}[htb]
\centering
\caption{Component ablation on nuScenes-SurroundOcc.}
\label{tab:component_ablation}
\scriptsize
\setlength{\tabcolsep}{3.5pt}
\renewcommand{\arraystretch}{1.15}
\resizebox{\columnwidth}{!}{
\begin{tabular}{l|c|ccc|cc}
\toprule
\textbf{Configuration}
& \textbf{\#B}
& \textbf{CGSS}
& \textbf{DRSS}
& \textbf{GSDF}
& \textbf{IoU} $\uparrow$
& \textbf{mIoU} $\uparrow$ \\
\midrule
GaussianFormer~\cite{huang2024gaussianformer} & 0 & & & & 29.83 & 19.10 \\
Single Mamba & 1 & & & & 30.05 & 19.18 \\
Dual Mamba$^\dagger$ & 2 & & & & 30.24 & 19.23 \\
+ GSDF & 2 & & & \cmark & 30.27 & 19.25 \\
+ CGSS & 2 & \cmark & & & 30.88 & 19.48 \\
+ DRSS & 2 & & \cmark & & 30.68 & 19.62 \\
+ CGSS + DRSS$^\dagger$ & 2 & \cmark & \cmark & & 31.82 & 20.25 \\
+ CGSS + GSDF & 2 & \cmark & & \cmark & 31.03 & 19.59 \\
+ DRSS + GSDF & 2 & & \cmark & \cmark & 30.91 & 19.72 \\
\rowcolor{lightbluegray}
\textbf{GaussianMamba (Ours)$^\dagger$}
& \textbf{2}
& \cmark
& \cmark
& \cmark
& \textbf{32.41}
& \textbf{20.55} \\
\bottomrule
\end{tabular}
}
\end{table}

\noindent\textbf{Branch capacity vs.\ scan order.}
Tables~\ref{tab:component_ablation} and~\ref{tab:ordering_ablation}
distinguish the effects of branch capacity and Gaussian
sequencing. Duplicating the Hilbert scan yields only
0.05 mIoU improvement, and replacing the duplicate with a reversed
(bidirectional) or an XYZ scan adds only 0.05 and 0.19 more.
With identical dual-branch capacity, CGSS and DRSS
outperform Hilbert\,+\,XYZ by 0.83 mIoU at 7.6\,ms
additional latency. Their joint improvement (1.02) also
exceeds the sum of their individual gains (0.25 and 0.39), suggesting
complementary geometric and semantic context.

\begin{table}[htb]
\centering
\caption{Comparison of Gaussian scanning strategies on nuScenes-SurroundOcc.
All dual-branch configurations employ independently parameterized
Mamba branches with concatenation-based fusion.}
\label{tab:ordering_ablation}
\scriptsize
\setlength{\tabcolsep}{4pt}
\renewcommand{\arraystretch}{1.15}
\resizebox{\columnwidth}{!}{
\begin{tabular}{l|c|ccc}
\toprule
\textbf{Scanning Strategy}
& \textbf{\#B}
& \textbf{IoU} $\uparrow$
& \textbf{mIoU} $\uparrow$
& \textbf{Latency (ms)} $\downarrow$ \\
\midrule
XYZ
& 1 & 29.96 & 19.06 & 6.99 \\
Morton & 1 & 30.00 & 19.13 & \textbf{6.10} \\
Hilbert & 1 & 30.05 & 19.18 & 10.68 \\
\midrule
Hilbert + Hilbert & 2 & 30.24 & 19.23 & 16.16 \\
Hilbert + Reversed Hilbert & 2 & 30.29 & 19.28 & 16.74 \\
Hilbert + XYZ & 2 & 30.48 & 19.42 & 17.40 \\
\rowcolor{lightbluegray}
\textbf{CGSS + DRSS}$^\dagger$ & \textbf{2}
& \textbf{31.82} & \textbf{20.25} & 24.98 \\
\bottomrule
\end{tabular}
}
\end{table}

\noindent\textbf{Gaussian support in CGSS.}
Table~\ref{tab:cgss_ablation} isolates the contribution
of anisotropic Gaussian support. Constrained reordering
without support improves mIoU by 0.29 while barely changing the share of
adjacent pairs in contact (12.6\% to 13.4\%), whereas incorporating support
raises contact to 18.7\% and adds a further 0.34 mIoU.
Both the constrained reordering itself and the Gaussian extent
beyond center proximity therefore contribute to spatial sequencing.

\begin{table}[htb]
\centering
\caption{Ablation of CGSS design choices on nuScenes-SurroundOcc.
The second branch uses DRSS. Contact denotes the
percentage of adjacent Gaussian pairs satisfying
$g_{ij}\leq 1$. Removing support excludes
$\log(1+g_{ij})$ from the adjacency cost.}
\label{tab:cgss_ablation}
\scriptsize
\setlength{\tabcolsep}{5pt}
\renewcommand{\arraystretch}{1.15}
\resizebox{\columnwidth}{!}{
\begin{tabular}{lccc}
\toprule
\textbf{Geometry-Branch Order}
& \textbf{Contact (\%)} $\uparrow$
& \textbf{IoU} $\uparrow$
& \textbf{mIoU} $\uparrow$ \\
\midrule
Hilbert & 12.60 & 30.68 & 19.62 \\
CGSS w/o Gaussian support & 13.40 & 31.21 & 19.91 \\
\rowcolor{lightbluegray}
\textbf{CGSS } & \textbf{18.70} & \textbf{31.82} & \textbf{20.25} \\
\bottomrule
\end{tabular}
}
\end{table}

\noindent\textbf{DRSS components.}
Table~\ref{tab:drss_ablation} shows that group routing
and boundary correction independently improve mIoU
by 0.42 and 0.25, respectively. Their combined gain
of 0.70 is close to the sum of the individual gains (0.67), indicating
largely additive contributions from semantic ordering and
support-aware boundary correction.

\begin{table}[htb]
\centering
\caption{Ablation of DRSS components on nuScenes-SurroundOcc. The first branch uses CGSS; the first row represents plain directional scanning.}
\label{tab:drss_ablation}
\scriptsize
\setlength{\tabcolsep}{6pt}
\renewcommand{\arraystretch}{1.08}
\resizebox{\columnwidth}{!}{
\begin{tabular}{cc|cc}
\toprule
\textbf{Group routing} & \textbf{Boundary correction} & \textbf{IoU} $\uparrow$ & \textbf{mIoU} $\uparrow$ \\
\midrule 
& & 30.98 & 19.55 \\
\cmark & & 31.39 & 19.97 \\
& \cmark & 31.20 & 19.80 \\
\rowcolor{lightbluegray}
\cmark & \cmark & \textbf{31.82} & \textbf{20.25} \\
\bottomrule
\end{tabular}
}
\end{table}

\noindent\textbf{Fusion strategy.}
Table~\ref{tab:fusion_ablation} shows that cross-attention
requires 4.7$\times$ the parameters of concatenation
for only 0.04 additional mIoU. In contrast, GSDF
improves mIoU by 0.30 with only 0.40K additional
parameters and 2.0\,ms latency overhead.
Its contribution increases from 0.02 mIoU with
duplicated Hilbert scans to 0.10-0.11 with one
specialized branch and 0.30 with both
(Table~\ref{tab:component_ablation}).
Although the smaller differences should be interpreted with caution,
this trend is consistent with GSDF exploiting the difference
between the two scans rather than added capacity.

\begin{table}[htb]
\centering
\caption{Comparison of feature fusion strategies for CGSS and DRSS on nuScenes-SurroundOcc.}
\label{tab:fusion_ablation}
\scriptsize
\setlength{\tabcolsep}{4pt}
\renewcommand{\arraystretch}{1.15}
\resizebox{\columnwidth}{!}{
\begin{tabular}{l|cccc}
\toprule
\textbf{Fusion Strategy}
& \textbf{Params (K)}
& \textbf{IoU} $\uparrow$
& \textbf{mIoU} $\uparrow$
& \textbf{Latency (ms)} $\downarrow$ \\
\midrule
Addition & \textbf{16.51} & 31.54 & 19.96 & \textbf{0.61} \\
Concatenation & 32.80 & 31.82 & 20.25 & 0.82 \\
Cross-attention & 155.32 & 32.01 & 20.29 & 9.13 \\
\rowcolor{lightbluegray} \textbf{GSDF }$^\dagger$ & 33.20 & \textbf{32.41} & \textbf{20.55} & 2.80 \\
\bottomrule
\end{tabular}
}
\end{table}

\noindent\textbf{Computational Efficiency.}
Table~\ref{tab:efficiency} examines the trade-off between Gaussian count, efficiency, and prediction accuracy. Latency and peak GPU memory are measured during inference using trained models. Reducing Gaussian count substantially decreases memory usage and inference latency. GaussianMamba adds only 0.8M parameters with negligible memory overhead, improving both IoU and mIoU at either scale. With GaussianFormer, the gains are larger at 144K Gaussians (+2.58 IoU, +1.45 mIoU) than at 25.6K (+0.88 IoU, +0.97 mIoU); this does not generalize across backbones, however, as GaussianFormer-2 gains 2.42 IoU and 1.10 mIoU with only 12.8K Gaussians. The additional scanning operations increase inference latency by 30.61 ms and 111.42 ms at 25.6K and 144K Gaussians, respectively; the latter is consistent with four scanning layers of roughly 27.8 ms each (Tables~\ref{tab:ordering_ablation} and~\ref{tab:fusion_ablation}), highlighting a computational trade-off, particularly for larger Gaussian sets.

\begin{table}[htb]
\centering
\caption{Computational efficiency and performance under different Gaussian counts on nuScenes-SurroundOcc.}
\label{tab:efficiency}
\scriptsize
\setlength{\tabcolsep}{3pt}
\renewcommand{\arraystretch}{1.08}
\resizebox{\columnwidth}{!}{
\begin{tabular}{l|c|ccccc}
\toprule
\textbf{Method} & \textbf{\#G} & \textbf{Params (M)} & \textbf{Mem. (GB)} $\downarrow$ & \textbf{Lat. (ms)} $\downarrow$ & \textbf{IoU} $\uparrow$ & \textbf{mIoU} $\uparrow$ \\
\midrule
GaussianFormer           & 144K & 75.55 & 2.77 & 354.71 & 29.83 & 19.10 \\
\rowcolor{lightbluegray} 
\textbf{GaussianFormer + Ours}$^\dagger$    & 144K & 76.35 & 2.79 & 466.12 & \textbf{32.41} & \textbf{20.55} \\
\midrule
GaussianFormer           & 25.6K & \textbf{57.09} & \textbf{2.08} & \textbf{183.51} & 28.72 & 16.00 \\
\rowcolor{lightbluegray} 
\textbf{GaussianFormer + Ours}$^\dagger$    & 25.6K & 57.89 & 2.11 & 214.12 & 29.60 & 16.97 \\
\bottomrule
\end{tabular}}
\end{table}

\section{Conclusion}
\label{sec:conclusion}

We presented GaussianMamba, a dual-branch Mamba framework that incorporates Gaussian-specific geometry and feature-semantic relationships into sequence modeling for 3D semantic occupancy prediction. CGSS accounts for anisotropic Gaussian support during spatial sequencing, DRSS captures feature-semantic relationships beyond geometric adjacency, and GSDF fuses the two scans through support-weighted scan-local discrepancies. On nuScenes-SurroundOcc and SSCBench-KITTI-360, GaussianMamba improves GaussianFormer and GaussianFormer-2 by 2.2-2.6 IoU and 0.9-1.5 mIoU points with modest parameter and memory overhead. The main limitation is increased inference latency, reaching approximately 31\% overhead at 144K Gaussians. Future work will focus on more efficient Gaussian scanning to reduce this computational cost.

\section*{LLM Usage Statement}
OpenAI's ChatGPT and Anthropic's Claude were used to assist with manuscript preparation, including grammar and spelling correction, language refinement, identification of typographical errors, and suggestions for improving clarity and consistency across the text. The authors reviewed the suggested revisions and remain responsible for the scientific content, technical claims, experimental results, and final manuscript.

{
    \small
    \bibliographystyle{ieeenat_fullname}
    \bibliography{main}
}

% WARNING: do not forget to delete the supplementary pages from your submission 
% \input{sec/X_suppl}

\end{document}